\documentclass{article}
\PassOptionsToPackage{table}{xcolor}
\usepackage{iclr2027,times}

\usepackage{amsmath,amsfonts,bm}

\def\eqref#1{equation~\ref{#1}}
\def\1{\bm{1}}

\DeclareMathAlphabet{\mathsfit}{\encodingdefault}{\sfdefault}{m}{sl}
\SetMathAlphabet{\mathsfit}{bold}{\encodingdefault}{\sfdefault}{bx}{n}

\usepackage{amsmath}
\usepackage{amssymb}
\usepackage{array}
\usepackage{booktabs}
\usepackage{caption}
\usepackage{enumitem}
\usepackage{graphicx}
\usepackage{hyperref}
\hypersetup{hidelinks}
\usepackage{multirow}
\usepackage{url}
\usepackage{wrapfig}
\usepackage{float}
\usepackage{placeins}
\usepackage{tikz}
\usetikzlibrary{arrows.meta,positioning}
\newcolumntype{L}[1]{>{\raggedright\arraybackslash}p{#1}}
\newcolumntype{C}[1]{>{\centering\arraybackslash}p{#1}}

\definecolor{occBike}{RGB}{244,160,170}
\definecolor{occBus}{RGB}{241,120,52}
\definecolor{occCar}{RGB}{31,144,224}
\definecolor{occMotor}{RGB}{188,165,0}
\definecolor{occPed}{RGB}{226,35,35}
\definecolor{occTruck}{RGB}{153,78,24}
\definecolor{occOther}{RGB}{125,125,125}
\definecolor{occBarrier}{RGB}{96,62,45}
\definecolor{occCone}{RGB}{245,213,93}
\definecolor{occDrive}{RGB}{230,30,225}
\definecolor{occSide}{RGB}{85,21,91}
\definecolor{occTerrain}{RGB}{125,224,80}
\definecolor{occManmade}{RGB}{196,196,196}
\definecolor{occVegetation}{RGB}{0,185,0}
\definecolor{occFree}{RGB}{0,0,0}

\newcommand{\classswatch}[1]{\textcolor{#1}{\rule{0.70em}{0.70em}}}
\newcommand{\occclassname}[1]{\rotatebox{90}{\scriptsize #1}}

\newcommand{\method}{RoadOcc}

\newcommand{\cmark}{\checkmark}

\definecolor{hitmatch}{RGB}{64,176,64}
\definecolor{hitmismatch}{RGB}{220,55,55}
\definecolor{hitother}{RGB}{160,160,160}
\DeclareRobustCommand{\hitbox}[1]{%
  \raisebox{0.12ex}{\textcolor{#1}{\rule{1.0ex}{1.0ex}}}}

\title{RoadOcc Learns When to Persist, Transport, or Refresh Memory for Roadside Occupancy Prediction}
\author{
Xiaokai Bai\textsuperscript{1}\textsuperscript{*},
Lei Yang\textsuperscript{2}\textsuperscript{*},
Songkai Wang\textsuperscript{1},
Lianqing Zheng\textsuperscript{3},
Siyuan Cao\textsuperscript{1},
Hui-liang Shen\textsuperscript{1}\textsuperscript{$\dagger$}
\\[2mm]
\textsuperscript{1}College of Information Science and Electronic Engineering, Zhejiang University.
\\
\textsuperscript{2}School of Mechanical and Aerospace Engineering, Nanyang Technological University.
\\
\textsuperscript{3}School of Automotive Studies, Tongji University.
\textsuperscript{*} Equal contribution, \textsuperscript{$\dagger$} Corresponding author.
\\
\texttt{email:shawnnnkb@gmail.com}
}
\iclrfinalcopy
\begin{document}

\maketitle

\begin{abstract}
Fixed roadside cameras repeatedly observe a stable scene overlaid by sparse
moving traffic. Temporal memory can recover weak observations, but reusing
moving evidence at stale locations can corrupt occupancy predictions. Motion
compensation addresses displacement, while reliance on the resulting history
remains a separate learning problem. We introduce RoadOcc, which learns
soft routing among fixed-coordinate history
(\emph{Persist}), velocity-addressed history (\emph{Transport}), and current
evidence (\emph{Refresh}). Motion state and class-consistent historical support
supervise these source choices. Dynamic-aware cross-attention (DCA) updates
candidate locations, multi-scale voxel velocity estimation (VVE) constructs
transport addresses from multi-scale current--history correspondence, and
velocity-guided dynamic sparse fusion (VDSF) combines
routed evidence under fixed sparse-token budgets. On InfraOcc, RoadOcc reaches
65.29 mIoU and 32.37 dynamic mIoU, gains of 4.44 and 4.71 over STCOcc.
Controlled address experiments show that VVE raises
dynamic mIoU by 0.87 over fixed-coordinate reading. Across three seeds,
supervised P/T/R adds 1.40 dynamic points over motion-corrected retrieval, while
removing Refresh costs 0.32 points. Results from two transfer models,
Occ3D-nuScenes, and longer intervals provide additional support. Code will be released.
\end{abstract}

\section{Introduction}
\label{sec:introduction}

Semantic occupancy prediction reconstructs free space, static layout, and
traffic participants in a common voxel volume. Camera-only methods have made
this representation practical through sparse voxel queries, BEV lifting,
tri-plane features, and unified panoptic volumes
\citep{MonoScene,VoxFormer,OpenOccupancy,Occ3D2023,BEVFormer,TPVFormer,
SurroundOcc,SparseOcc,GaussianFormer}. However, these methods
are predominantly developed for moving platforms, where each observation is
interpreted in an ego-centric frame. A roadside sensor instead repeatedly
observes a persistent scaffold overlaid by sparse, short-lived dynamics. History
is most valuable for weak or occluded actors, yet stale footprints can overwrite
exactly these targets. The fixed world frame therefore shifts the central
temporal problem to local evidence management.
Static history is already aligned, whereas dynamic history must be selectively
relocated or rejected.

The same history can provide either valid support or stale evidence.
Fixed-coordinate memory preserves static layout and can recover weak or occluded vehicle
boundaries, but it does not relocate an object that moved within the fixed
coordinate system. Reading such memory from the former footprint can blur a
dynamic boundary, duplicate an object, or attach it to nearby static layout.
Discarding history avoids these errors but also removes useful support. The
challenge is therefore not simply whether to use history, but to decide where
temporal evidence is needed, where supporting history should be read, and
how strongly to rely on each source. These cases can coexist within one scene
or along a single object boundary, making a uniform temporal fusion rule
insufficient.

Recent temporal methods demonstrate the value of multi-frame context for
occupancy, flow, and 3D perception \citep{Cam4DOcc,OccWorld,UniOcc,ALOcc,
STCOcc,LetOccFlow,BEVDet4D,RecurrentBEV,STOccMemory}. Existing designs
already incorporate temporal memory, motion alignment, and adaptive fusion.
Motion alignment addresses displacement, but a flow-proposed
address may still lack class-consistent support for the current voxel. A
stationary object may reuse fixed-coordinate memory, a moving object may retrieve
memory near $\mathbf x-\Delta t\,\mathbf v_t(\mathbf x)$, and a newly appearing
object may have to rely on current evidence. Adaptive weighting can represent
these alternatives. Our focus is to supervise the source choice explicitly
using motion state and semantic support, giving temporal fusion a structured
learning target.
\begin{wrapfigure}{r}{0.65\columnwidth}
  \centering
  \includegraphics[width=0.65\textwidth]{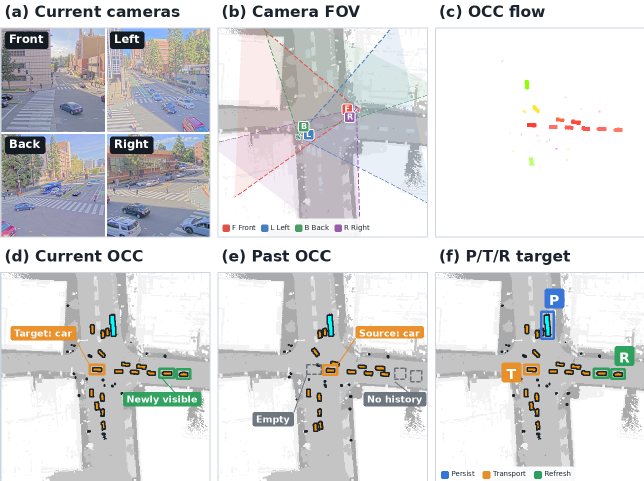}
  \caption{\textbf{Real-scene evidence for explicit source routing.}
  The top row shows current views, camera FOVs, and current-aligned occupancy
  flow (a--c). The bottom row shows current/past occupancy and P/T/R targets
  (d--f), illustrating a
  persistent bus, transported car, and newly visible cars assigned Refresh.}
  \label{fig:motivation}
  \vspace{-10pt}
\end{wrapfigure}
We therefore formulate temporal occupancy as explicit evidence-source routing.
\emph{Should each selected voxel Persist, Transport, or Refresh?} Persist reuses
the fixed-coordinate source, Transport applies backwarp sampling to
velocity-addressed history, and Refresh uses no history, only current evidence.
Flow proposes where history should be read, while P/T/R learns preferences
over the three sources. The supervision favors supported fixed history for
stationary targets and supported transported history for moving targets,
using Refresh when the corresponding candidate lacks support.

RoadOcc realizes this principle through a DCA--VVE--VDSF chain. DCA
localizes the dynamic candidate and refines the current observation. VVE then
combines current and temporal flow into dynamic flow, after which VDSF selects
fixed-coordinate, backwarped, or current evidence. The stages therefore resolve
where temporal reasoning is needed, where history should be read, and which
source should enter fusion. Their intermediate outputs allow separate
diagnostics of dynamic localization, historical addressing, and source selection.
Figure~\ref{fig:motivation} grounds the three evidence-source alternatives. 
Our contributions are summarized below.

\begin{itemize}[leftmargin=10pt]
  \item We propose \method{}, which formulates roadside temporal occupancy as
  a Persist--Transport--Refresh source-selection problem. Motion state and
  class-consistent support supervise the source preferences, complementing
  motion-based historical addressing.
  \item We implement this formulation in a sparse, multi-scale pipeline.
  DCA allocates image updates, VVE estimates transport addresses, and VDSF
  combines routed sources while retaining a short background memory path.
  \item We report joint occupancy and motion improvements on InfraOcc,
  three-seed routing controls, and module transfer to two temporal models.
  Occ3D-nuScenes and temporal-gap evaluations further test the design beyond
  the primary roadside setting and training cadence.
\end{itemize}

\section{Related Work}
\label{sec:related}

\subsection{Camera Occupancy and Roadside Perception}
Camera occupancy methods lift multi-view image evidence into volumetric
representations. MonoScene \cite{MonoScene} and VoxFormer \cite{VoxFormer} establish image-based semantic scene
completion, while BEVDet, BEVDepth, and BEVFormer improve multi-view lifting and
BEV reasoning \citep{BEVDet,BEVDepth,BEVFormer}. Occupancy-specific designs
span tri-plane, dense,
sparse, Gaussian, and forward--backward representations
\citep{TPVFormer,SurroundOcc,OccFormer,SparseOcc,GaussianFormer,FBOCC}, as well
as panoptic and instance-aware prediction
\citep{PanoOcc,VoxDet}. Unlike MC-BEVRO's roadside BEV occupancy monitoring
\citep{MCBEVRO}, RoadOcc routes evidence for class-aware 3D occupancy.
\begin{wrapfigure}{r}{0.65\columnwidth}
  \vspace{-5pt}
  \centering
  \includegraphics[width=\linewidth]{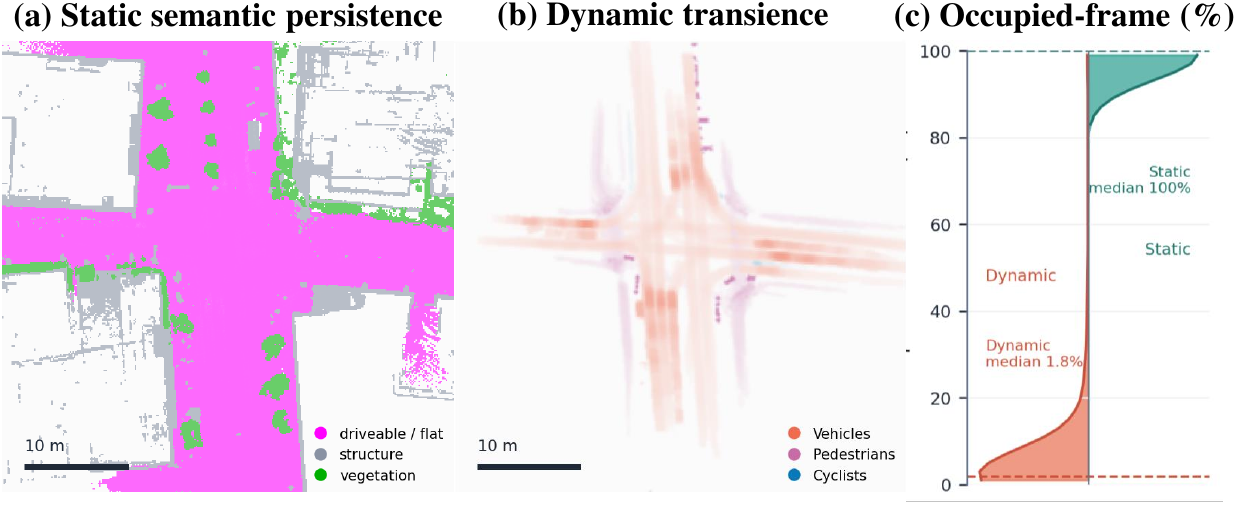}
  \vspace{-15pt}
  \caption{\textbf{Static-to-dynamic persistence in InfraOcc.}}
  \label{fig:datasetvis}
  % \vspace{-13pt}
\end{wrapfigure}
\vspace{-1pt}
\noindent InfraOcc provides a roadside occupancy benchmark
emphasizing persistent infrastructure \citep{InfraOcc}.
Figure~\ref{fig:datasetvis} reveals its temporal asymmetry. Fixed roadway
structure provides broad, stable support, while dynamic participants form
sparse, short-lived trails. This asymmetry favors retaining static context while
selectively relocating moving evidence.

\subsection{Temporal Occupancy, Motion, and Alignment}

Temporal occupancy methods improve memory through recurrent aggregation,
motion-aware alignment, and explicit association. Recurrent and video-based
models maintain or forecast scene state
\citep{BEVDet4D,RecurrentBEV,SelfOcc,ForecastOcc}, while occupancy--flow methods
couple volumetric semantics with motion
\citep{Cam4DOcc,OccWorld,UniOcc,ALOcc,STCOcc,LetOccFlow,OFMPNet,OmniHDScenes}.
Other approaches organize temporal cues, learn semantic associations, or
retrieve traversal priors \citep{STOccMemory,GDFusion,LinkOcc,LMPOcc}.
RoadOcc builds on temporal alignment and adaptive fusion by supervising a
choice among fixed-coordinate history, transported history, and current
evidence. Its emphasis is the motion- and support-based learning target for
source selection, rather than the weighted-sum operation alone.

\begin{table*}[htbp]
  \centering
  \begin{minipage}[t]{0.65\textwidth}
    \centering
    \caption{\textbf{Generated GT-flow target at $t+0.5$ s.}
    Zero-flow and GT-flow transport share the inter-frame transform and differ
    only by object displacement. Same/Dyn./Occ. denote unchanged semantics,
    dynamic class, and occupied class. Values are percentages and $N$
    counts valid queries in millions.}
    \label{tab:temporal_verification}
    \footnotesize
    \setlength{\tabcolsep}{0pt}
    \setlength{\belowrulesep}{0pt}
    \setlength{\aboverulesep}{0pt}
    \begin{tabular}{@{}L{2.03cm}C{0.80cm}C{0.85cm}C{0.85cm}C{0.85cm}C{0.85cm}C{0.85cm}C{0.85cm}C{1.08cm}@{}}
      \toprule[1.0pt]
      \multirow{2}{=}{Speed} & \multirow{2}{=}{$N$ (M)} &
      \multicolumn{3}{C{2.55cm}}{Zero-flow transport} &
      \multicolumn{3}{C{2.55cm}}{GT-flow warp} &
      \multirow{2}{=}{$\Delta$ Same} \\
      \cmidrule(lr){3-5}\cmidrule(lr){6-8}
      & & Same & Dyn. & Occ. & Same & Dyn. & Occ. & \\
      \midrule
      All & 11.83 & 72.93 & 78.86 & 90.92 & \textbf{89.21} & \textbf{95.79} & \textbf{99.38} & $+$16.28 \\
      \midrule
      Slow ($<2$ m/s) & 7.87 & 85.19 & 94.08 & 98.80 & 86.51 & \textbf{96.25} & \textbf{99.70} & $+$1.32 \\
      Med. (2--5 m/s) & 1.16 & 71.08 & 71.38 & 87.50 & \textbf{96.65} & \textbf{97.64} & \textbf{99.53} & $+$25.57 \\
      Fast ($\geq5$ m/s) & 2.80 & 39.57 & 39.59 & 70.41 & \textbf{93.73} & \textbf{93.75} & \textbf{98.43} & $+$54.16 \\
      \bottomrule[1.0pt]
    \end{tabular}
  \end{minipage}\hfill
  \begin{minipage}[t]{0.34\textwidth}
    \centering
    \captionof{figure}{\textbf{Motion compensation restores temporal consistency.}
     GT flow preserves dynamic support across speed ranges, whereas fixed-coordinate
     retrieval becomes increasingly misaligned as speed grows.}
     \vspace{-5pt}
    \includegraphics[width=\linewidth]{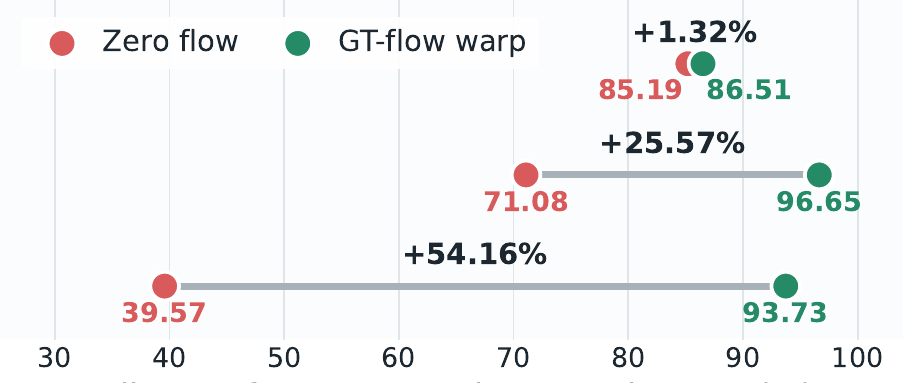}
    \label{fig:speed_temporal_alignment}
  \end{minipage}
\end{table*}
\vspace{-10pt}

Flow-based alignment is closest to this problem. AsyncBEV aligns asynchronous
cross-modal observations with BEV feature flow \citep{AsyncBEV}. In the fixed
roadside setting, Table~\ref{tab:temporal_verification} and
Figure~\ref{fig:speed_temporal_alignment} show that GT-flow warping improves
same-class agreement by 1.32 points for slow objects, but by 25.57 and
54.16 points for medium and fast objects, motivating selective motion
compensation. This is a diagnostic of the constructed targets. For candidates
no faster than 2\,m/s, low-speed target repair selects zero motion when its
future same-class support score exceeds the transported score by at least 0.05.

% Dynamic voxels are sparse and affected by class imbalance and occlusion. Focal
% and class-balanced objectives address related long-tail effects
% \citep{FocalLoss,ClassBalancedLoss}. However, reweighting rare classes does not
% resolve their heterogeneous temporal states because historical evidence may already be
% aligned, displaced, or unsupported. Class imbalance and temporal provenance are
% therefore complementary problems.

\section{Method}
\label{sec:method}

\subsection{Preliminary Analysis}

Let $\mathcal I_t=\{I_t^n\}_{n=1}^{4}$ denote the synchronized roadside images
at time $t$. RoadOcc predicts semantic occupancy $\mathbf y_t$ over a voxel
domain $\Omega$ and a planar velocity field
$\mathbf v_t:\Omega\rightarrow\mathbb R^2$. Among the 18 occupancy states, the
dynamic set is $\mathcal C_{\rm dyn}=\{\text{bicycle, bus, car, motorcycle,
pedestrian, truck}\}$. Velocity is expressed in the common roadside horizontal
frame.

Because fixed roadside cameras share a persistent physical frame, cached
history already lies on the current voxel grid at inference. We only reconcile
relative BEV augmentations during training. Under a locally constant velocity
approximation, history supporting a current voxel
$\mathbf x$ is read from
\begin{equation}
  \widetilde{\mathbf h}_{t-\Delta t}(\mathbf x)=
  \operatorname{Sample}\!\left(\overline{\mathbf h}_{t-\Delta t},
  \mathbf x-\Delta t\,\mathbf v_t(\mathbf x)\right),
  \label{eq:warp}
\end{equation}
where $\overline{\mathbf h}$ is same-grid history and
$\operatorname{Sample}$ is trilinear interpolation. The three candidates are
fixed-coordinate history at $\mathbf x$ (Persist), flow-addressed history at
$\mathbf x-\Delta t\,\mathbf v_t(\mathbf x)$ (Transport), and current evidence
(Refresh). Equation~\ref{eq:warp} specifies an address, while VDSF learns how to
combine these sources under motion- and support-based supervision.

\subsection{Overview}
\label{sec:method_overview}

\begin{figure*}[t]
  \centering
  \includegraphics[width=\textwidth]{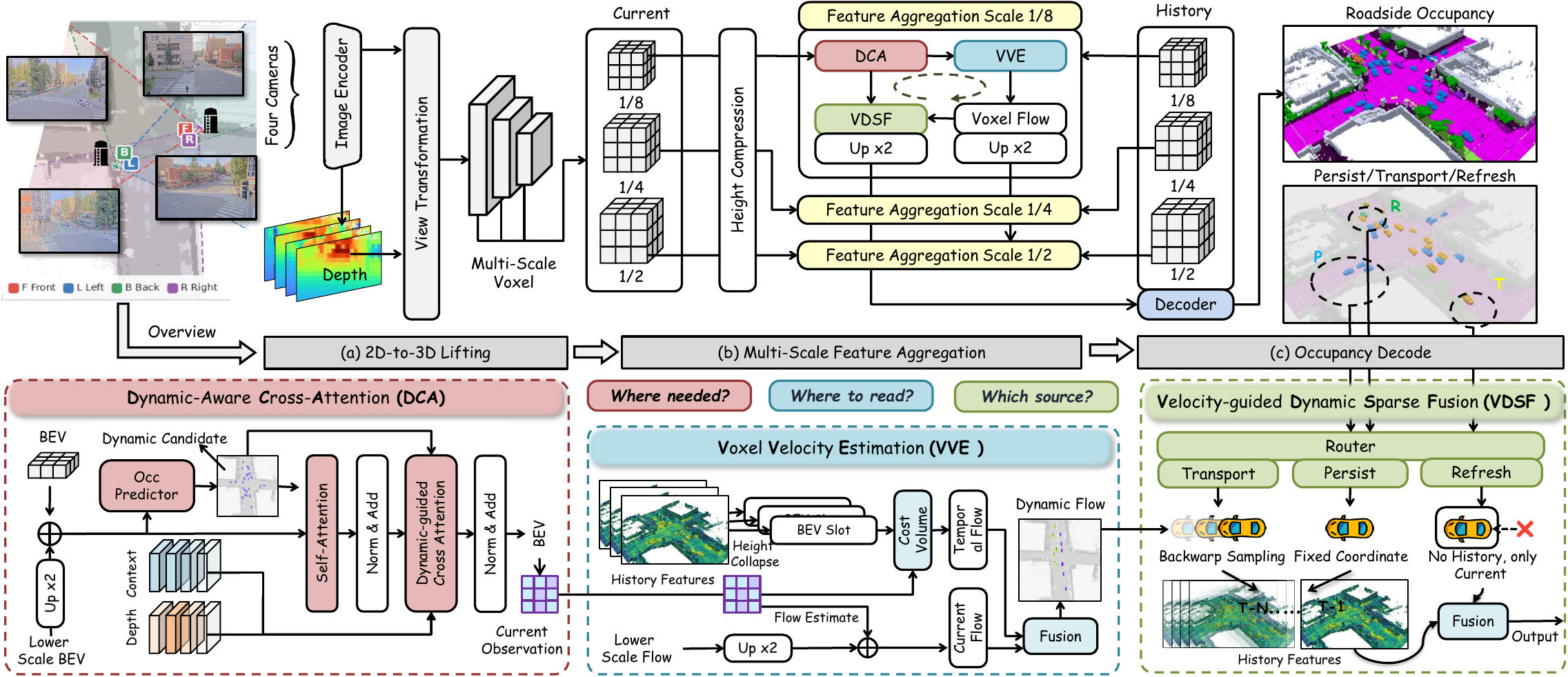}
  \caption{\textbf{RoadOcc framework.} The image encoder and view
  transformation construct multi-scale voxel features. At each feature
  aggregation scale, DCA produces the current observation, VVE fuses current,
  temporal, and lower-scale flow into dynamic flow, and VDSF routes
  fixed-coordinate, backwarped, or current evidence to fusion. Coarse-to-fine
  outputs feed occupancy decoding for semantic occupancy and voxel flow.}
  \label{fig:overview}
\end{figure*}

RoadOcc decodes $\{\mathbf V_t^s\}_{s\in\{1/8,1/4,1/2\}}$ from coarse to fine.
As shown in Figure~\ref{fig:overview}, the encoder lifts four camera views into
multi-scale voxel features. At each aggregation stage, DCA determines where
evidence is needed, VVE determines where history should be read, and VDSF
determines which source should enter fusion. The output feature and dynamic flow
seed the next finer stage before occupancy decoding. Coarse predictions
establish scene layout, while finer stages restore object boundaries and sparse
traffic participants before the full-resolution output.

\subsection{Dynamic-aware Cross-attention (DCA)}
\label{sec:dca}

Dynamic objects occupy few voxels, so uniformly repeating image cross-attention
spends most computation on stable structure. After the first nonempty-guided
image update, DCA combines the current representation with an upsampled
lower-scale BEV to derive a dynamic candidate. Let
$\boldsymbol\pi_{t,\rm cur}^s$ and $\boldsymbol\pi_{t,\rm sta}^s$ denote the
provisional distributions produced by the current and static prediction heads
after the first image update, and let
$\mathcal C_{\rm sta}^{+}$ contain the static classes and free space. The mean
semantic discrepancy $\delta_{t,\rm sta}^s$ over $\mathcal C_{\rm sta}^{+}$ and
current dynamic probability $p_{t,\rm dyn}^s$ form the candidate map
\begin{equation}
  c_t^s(\mathbf x)
  =\operatorname{Norm}\!\left(\max\!\left\{
  \delta_{t,\rm sta}^s(\mathbf x),p_{t,\rm dyn}^s(\mathbf x)
  \right\}\right).
  \label{eq:dca_candidate}
\end{equation}
Here $\operatorname{Norm}$ divides by the per-sample spatial maximum, clamped to
$10^{-6}$. Discrepancy recovers departures from the static hypothesis, while
dynamic probability retains visible actors. Their maximum also preserves early
semantic misses detected through change evidence.

The candidate map controls a second image cross-attention rather than directly
reweighting the voxel feature. Following projected deformable attention
\citep{BEVFormer} and occupancy-aware reference sampling
\citep{STCOcc}, let $\mathbf q_{t,\mathbf x}^s$ be the BEV query at
$\mathbf x$, $\mathcal R_n(\mathbf x)$ its visible height anchors in camera
$n$, and $\mathbf F_t^n$ the corresponding image feature. DCA retains
$\mathcal R_{n,c}^s(\mathbf x)=\{\mathbf r\in\mathcal R_n(\mathbf x):
u_{\mathbf r}<c_t^s(\mathbf r)\}$ and updates the query as
\begin{equation}
 \begin{aligned}
 \mathbf q_{t,\mathbf x}^{s,+}=\mathbf q_{t,\mathbf x}^{s}
  +\mathbf W_o\frac{1}{|\mathcal V_{\mathbf x}|}
  \sum_{n\in\mathcal V_{\mathbf x}}\sum_{\mathbf r\in
  \mathcal R_{n,c}^s(\mathbf x)}
  \beta_{n,\mathbf r}^s\,
  \mathcal A_{\rm def}\!\left(
  \mathbf q_{t,\mathbf x}^{s},\mathcal P_n(\mathbf r),\mathbf F_t^n\right),
 \end{aligned}
 \label{eq:dca_cross_attention}
\end{equation}
where $\mathbf r=(\mathbf x,z_{\mathbf r})$ indexes a voxel-height score shared
across cameras, $\mathcal V_{\mathbf x}$ contains valid cameras, and
$\mathcal P_n$ projects anchors. Depth consistency $\beta_{n,\mathbf r}^s$
multiplies normalized deformable weights before camera averaging. During
training, $u_{\mathbf r}$ is sampled from a normal distribution
with mean $0.5$ and standard deviation $1$, truncated to $[0,1]$. At inference,
it is fixed to $0.5$. The stochastic
rule exposes uncertain boundary voxels during training, while deterministic
thresholding keeps inference stable. Queries without retained support follow
the residual path unchanged.

The candidate gates a second cross-attention over image depth context and also
allocates VDSF tokens. DCA therefore localizes where additional image and
temporal evidence is needed without repeatedly updating the stable scene.

\subsection{Multi-scale Voxel Velocity Estimation (VVE)}
\label{sec:vve}

DCA refines the current observation but does not locate displaced history. VVE
estimates its address from the DCA feature and nearest valid history. From the
current occupancy distribution $\boldsymbol\pi_t^s(\mathbf x)$, it computes
dynamic support
$d_t^s=\sum_{c\in\mathcal C_{\rm dyn}}\pi_{t,c}^s$ and nonempty support
$e_t^s=1-\pi_{t,\rm free}^s$. Here $d_t^s$ guides height compression into the
current BEV slot $\mathbf B_t^s$, while $e_t^s$ guides temporal token selection.
Applying the same compression to history gives
$\overline{\mathbf B}_{t-\Delta t}^s$.

VVE next compares each current BEV slot with a small neighborhood in the
historical slot, producing the local cost volume
\begin{equation}
  \mathbf K_t^s(\mathbf x,\delta)=
  \left\langle
    \frac{\mathbf B_t^s(\mathbf x)}{\|\mathbf B_t^s(\mathbf x)\|_2},
    \frac{\overline{\mathbf B}_{t-\Delta t}^s(\mathbf x+\delta)}
         {\|\overline{\mathbf B}_{t-\Delta t}^s(\mathbf x+\delta)\|_2}
  \right\rangle,
  \quad \delta\in[-r,r]^2,
  \label{eq:cost_volume}
\end{equation}
where $r=2$ yields a $5\times5$ neighborhood and both norms use an
$\epsilon=10^{-6}$ floor. The cost-volume decoder converts this local
correspondence evidence into temporal flow.

VVE estimates flow recursively rather than solving each resolution in
isolation. Let $s^-$ denote the preceding coarser stage and set the upsampled
lower-scale flow $\mathbf v_{t,\uparrow}^s$ to zero at $1/8$ and to
$\operatorname{Up}_s(\mathbf v_{t,\rm cur}^{s^-})$ otherwise. This prior
initializes the current flow. Current appearance and its semantic embedding
$\mathbf S_t^s$ predict a
current-only correction, while the history feature and cost volume predict the
temporal flow. Their fusion is
\begin{equation}
 \begin{aligned}
  \mathbf v_{t,\rm cur}^s
    &=\mathbf v_{t,\uparrow}^s+\Delta\mathbf v_{t,\rm cur}^s,
  \quad \mathbf v_{t,\rm tmp}^s
    =f_{\rm tmp}(\mathbf B_t^s,\overline{\mathbf B}_{t-\Delta t}^s,
      \mathbf K_t^s),
      \\\widehat{\mathbf v}_t^s
    &=(1-\alpha_t^s)\mathbf v_{t,\rm cur}^s
      +\alpha_t^s\mathbf v_{t,\rm tmp}^s
      +\Delta\mathbf v_{t,\rm ref}^s.
 \end{aligned}
  \label{eq:velocity_prior}
\end{equation}
Here $f_{\rm tmp}$ is the cost-volume decoder and
$\alpha_t^s=0.5\,\mathbf 1_{\rm hist}\max_z d_t^s$ gates history, becoming zero
at sequence starts. The residual $\Delta\mathbf v_{t,\rm ref}^s$ produces the
dynamic flow, which is lifted across height as the VDSF address. Coarse stages
capture motion structure and fine stages refine boundaries.

\subsection{Velocity-guided dynamic sparse fusion (VDSF)}
\label{sec:vdsf}

\paragraph{Velocity-guided sparse fusion.}
Applying the full temporal memory after DCA and VVE would mostly repeat static
context. VDSF makes gather--fuse--scatter \citep{STCOcc} motion aware by
combining DCA guidance with nonempty support under a fixed token budget
\begin{equation}
  \mathbf g_t^s=\operatorname{clip}
  (c_t^s+\eta\,e_t^s,0,1),
  \qquad
  \mathcal X_t^s=\operatorname{TopK}(\mathbf g_t^s),
  \label{eq:vdsf_selection}
\end{equation}
where $\eta$ balances change evidence and occupied support. The selected set
$\mathcal X_t^s$ receives full stage-local history. Its complement uses a short
fixed-coordinate path, preserving background context at lower cost.

Each scale receives multi-frame history features and their VVE motion priors.
For $\mathbf x\in\mathcal X_t^s$, elapsed time converts dynamic flow into a
voxel offset. Backwarp sampling produces the Transport candidate $\mathbf h^T$,
while the fixed-coordinate read produces the Persist candidate $\mathbf h^P$.

\paragraph{Persist--Transport--Refresh routing.}
Dynamic flow proposes where to read history, but a corrected address may still
lack semantic support. VDSF therefore decides which source should
enter fusion. It retains $\mathbf h^P$ and $\mathbf h^T$ and defines the
current-only Refresh candidate as $\mathbf h^R=\mathbf q_t$. During training,
the route target
is constructed from the current and nearest historical semantic occupancy
labels together with the motion target, and assigns each current dynamic voxel
according to its motion state and class-consistent historical support as follows.
\begin{equation}
 z_t(\mathbf x)=
 \begin{cases}
 P, & \|\mathbf v_t(\mathbf x)\|\leq\epsilon
      \ \land\ \operatorname{Supp}(\mathbf x),\\
 T, & \|\mathbf v_t(\mathbf x)\|>\epsilon
      \ \land\ \operatorname{Supp}
      (\mathbf x-\Delta t\,\mathbf v_t(\mathbf x)),\\
 R, & \text{otherwise},
 \end{cases}
 \label{eq:ptr_target}
\end{equation}
where $\operatorname{Supp}$ tests same-class occupancy at the same height within
the local correspondence radius. We use $\epsilon=10^{-3}$ m/s and a one-voxel
radius to absorb discretization error. The construction follows the semantic
occupancy label space used by InfraOcc and is evaluated at every execution
scale using its physical cell size, with a native-grid endpoint for
full-resolution supervision. The motion target selects which historical
candidate is tested according to motion state. The fixed-coordinate candidate
is used for stationary voxels and the transported candidate for moving voxels.
If that candidate lacks support, the target is Refresh,
even if the alternative historical address has same-class support. This is a
motion-conditioned source preference. Targets use GT motion, while inference uses
predicted VVE motion and learned route probabilities. Future occupancy is used
only for offline target construction, never as model input.

At each feature aggregation scale, the router combines the current observation,
nearest fixed-coordinate history feature, normalized dynamic flow, and
upsampled coarser P/T/R distribution. It predicts stage-local logits. A
lightweight native head upsamples the finest context with full-resolution flow.
Softmax gives
$\boldsymbol\alpha_t^s=(\alpha_P,\alpha_T,\alpha_R)$ and the routed candidate
\begin{equation}
 \mathbf h^{\rm PTR}_t=\alpha_P\mathbf h^P+
 \alpha_T\mathbf h^T+\alpha_R\mathbf h^R,\qquad
 (\alpha_P,\alpha_T,\alpha_R)=\operatorname{softmax}(\psi(\cdot)).
 \label{eq:ptr_route}
\end{equation}
P/T/R therefore compete as one route distribution rather than independent
weights. Transport uses backwarp sampling, Persist uses the fixed-coordinate
source, and Refresh uses no history, only current evidence. Fusion uses soft
probabilities and reserves argmax for diagnostics.

% Defined here to keep the full-width class table close to its discussion in
% Sec.~\ref{sec:main_results}, rather than allowing it to split the objective.
\newcommand{\mainresultstable}{%
\begin{table}[t]
  \caption{\textbf{Main InfraOcc occupancy results.}}
  %RoadOcc (ours) uses
  % strictly current-frame-only input. The two rows marked $^{\mathrm T}$ report
  % their strongest evaluated multi-frame checkpoints. All remaining rows use
  % current-frame-only input.
  \label{tab:infraocc_main_results}
  \centering
  \definecolor{best}{RGB}{255, 180, 180}
  \definecolor{second}{RGB}{255, 230, 180}
  \definecolor{third}{RGB}{255, 255, 200}
  \newlength{\mainmethodwd}
  \newlength{\mainmetricwd}
  \newlength{\mainclasswd}
  \setlength{\mainmethodwd}{0.16\textwidth}
  \setlength{\mainmetricwd}{0.043\textwidth}
  \setlength{\mainclasswd}{\dimexpr(\textwidth-\mainmethodwd-4\mainmetricwd-2\arrayrulewidth)/15\relax}
  \newcommand{\methodcell}[1]{\makebox[\mainmethodwd][l]{##1}}
  \newcommand{\metriccell}[1]{\makebox[\mainmetricwd][c]{##1}}
  \newcommand{\classcell}[1]{\makebox[\mainclasswd][c]{##1}}
  \newcommand{\bestmetriccell}[1]{\cellcolor{best}\metriccell{\textbf{##1}}}
  \newcommand{\secondmetriccell}[1]{\cellcolor{second}\metriccell{\underline{##1}}}
  \newcommand{\thirdmetriccell}[1]{\cellcolor{third}\metriccell{##1}}
  \newcommand{\bestclasscell}[1]{\cellcolor{best}\classcell{\textbf{##1}}}
  \newcommand{\secondclasscell}[1]{\cellcolor{second}\classcell{\underline{##1}}}
  \newcommand{\thirdclasscell}[1]{\cellcolor{third}\classcell{##1}}
  \scriptsize
  \setlength{\tabcolsep}{1.5pt}
  \renewcommand{\arraystretch}{1.05}
  \resizebox{\textwidth}{!}{%
  \begin{tabular}{@{}c|c|*{3}{c}|*{15}{c}@{}}
  \toprule
  \methodcell{Method} & \metriccell{gIoU} & \multicolumn{3}{c|}{\metriccell{mIoU}} & \classcell{\occclassname{Bike}} & \classcell{\occclassname{Bus}} & \classcell{\occclassname{Car}} & \classcell{\occclassname{Motorcycle}} & \classcell{\occclassname{Pedestrian}} & \classcell{\occclassname{Truck}} & \classcell{\occclassname{Other}} & \classcell{\occclassname{Barrier}} & \classcell{\occclassname{Cone}} & \classcell{\occclassname{Driveable}} & \classcell{\occclassname{Sidewalk}} & \classcell{\occclassname{Terrain}} & \classcell{\occclassname{Manmade}} & \classcell{\occclassname{Vegetation}} & \classcell{\occclassname{Free}} \\
  \cmidrule(lr){3-5}
  \methodcell{} & \metriccell{} & \metriccell{all} & \metriccell{dyn} & \metriccell{sta} & \classcell{\classswatch{occBike}} & \classcell{\classswatch{occBus}} & \classcell{\classswatch{occCar}} & \classcell{\classswatch{occMotor}} & \classcell{\classswatch{occPed}} & \classcell{\classswatch{occTruck}} & \classcell{\classswatch{occOther}} & \classcell{\classswatch{occBarrier}} & \classcell{\classswatch{occCone}} & \classcell{\classswatch{occDrive}} & \classcell{\classswatch{occSide}} & \classcell{\classswatch{occTerrain}} & \classcell{\classswatch{occManmade}} & \classcell{\classswatch{occVegetation}} & \classcell{\classswatch{occFree}} \\
  \midrule
  \methodcell{BEVDet~'21} & \metriccell{68.62} & \metriccell{45.14} & \metriccell{21.19} & \metriccell{63.11} & \classcell{9.23} & \classcell{42.75} & \classcell{38.04} & \classcell{9.68} & \classcell{15.01} & \secondclasscell{12.45} & \classcell{42.88} & \classcell{70.36} & \classcell{53.77} & \classcell{72.04} & \classcell{68.07} & \classcell{73.09} & \classcell{64.28} & \classcell{60.38} & \classcell{94.87} \\
  \methodcell{BEVFormer~'22} & \metriccell{81.72} & \metriccell{47.76} & \metriccell{10.73} & \metriccell{75.53} & \classcell{7.97} & \classcell{20.39} & \classcell{17.65} & \classcell{3.47} & \classcell{8.62} & \classcell{6.30} & \classcell{42.47} & \thirdclasscell{76.26} & \classcell{72.83} & \classcell{85.19} & \classcell{81.95} & \classcell{88.95} & \classcell{79.52} & \classcell{77.05} & \classcell{97.56} \\
  \methodcell{BEVDepth~'23} & \metriccell{71.89} & \metriccell{46.81} & \metriccell{19.64} & \metriccell{67.19} & \classcell{14.84} & \classcell{33.87} & \classcell{38.74} & \classcell{3.42} & \classcell{16.86} & \classcell{10.14} & \classcell{48.73} & \classcell{71.05} & \classcell{62.37} & \classcell{73.31} & \classcell{69.32} & \classcell{76.07} & \classcell{70.51} & \classcell{66.15} & \classcell{95.54} \\
  \methodcell{TPVFormer~'23} & \metriccell{73.80} & \metriccell{43.09} & \metriccell{4.36} & \metriccell{72.13} & \classcell{0.10} & \classcell{3.69} & \classcell{13.80} & \classcell{0.41} & \classcell{6.86} & \classcell{1.33} & \classcell{42.86} & \classcell{75.29} & \classcell{69.69} & \classcell{80.61} & \classcell{77.62} & \classcell{86.63} & \classcell{72.61} & \classcell{71.70} & \classcell{96.12} \\
  \methodcell{SurroundOcc~'23} & \metriccell{83.78} & \metriccell{50.77} & \metriccell{11.37} & \metriccell{80.32} & \classcell{12.66} & \classcell{21.90} & \classcell{18.54} & \classcell{2.06} & \classcell{9.50} & \classcell{3.55} & \classcell{65.09} & \bestclasscell{79.04} & \classcell{72.70} & \classcell{86.38} & \classcell{83.99} & \classcell{91.10} & \classcell{83.16} & \classcell{81.09} & \classcell{97.80} \\
  \methodcell{CONet~'23} & \metriccell{78.56} & \metriccell{49.25} & \metriccell{18.93} & \metriccell{72.00} & \classcell{3.06} & \classcell{37.24} & \classcell{39.47} & \classcell{4.42} & \classcell{17.64} & \classcell{11.72} & \classcell{45.01} & \classcell{75.31} & \classcell{64.94} & \classcell{84.29} & \classcell{79.45} & \classcell{84.27} & \classcell{72.96} & \classcell{69.79} & \classcell{97.08} \\
  \methodcell{ALOcc~'25} & \metriccell{75.66} & \metriccell{43.31} & \metriccell{5.86} & \metriccell{71.40} & \classcell{8.41} & \classcell{4.04} & \classcell{8.52} & \classcell{0.00} & \classcell{4.48} & \classcell{9.71} & \classcell{63.90} & \classcell{68.97} & \classcell{63.40} & \classcell{90.09} & \classcell{77.77} & \classcell{78.16} & \classcell{65.07} & \classcell{63.85} & \classcell{97.12} \\
  \methodcell{Let Occ Flow~'24} & \metriccell{87.18} & \metriccell{51.38} & \metriccell{8.78} & \thirdmetriccell{83.34} & \classcell{11.27} & \classcell{11.53} & \classcell{14.07} & \classcell{4.34} & \classcell{8.12} & \classcell{3.34} & \classcell{62.06} & \classcell{75.06} & \secondclasscell{85.56} & \classcell{92.55} & \thirdclasscell{85.63} & \thirdclasscell{92.98} & \thirdclasscell{87.10} & \thirdclasscell{85.74} & \classcell{98.56} \\
  \methodcell{SparseOcc~'24} & \metriccell{71.61} & \metriccell{45.42} & \metriccell{9.41} & \metriccell{72.42} & \classcell{1.50} & \classcell{20.94} & \classcell{24.87} & \classcell{0.00} & \classcell{4.60} & \classcell{4.57} & \thirdclasscell{67.68} & \secondclasscell{76.37} & \classcell{61.51} & \classcell{74.67} & \classcell{76.59} & \classcell{88.54} & \classcell{68.79} & \classcell{65.19} & \classcell{96.31} \\
  \methodcell{GaussianFormer~'24} & \metriccell{77.89} & \metriccell{43.63} & \metriccell{10.45} & \metriccell{68.51} & \classcell{3.13} & \classcell{22.11} & \classcell{23.17} & \classcell{1.29} & \classcell{8.21} & \classcell{4.80} & \classcell{36.70} & \classcell{68.54} & \classcell{55.60} & \classcell{82.59} & \classcell{79.37} & \classcell{74.20} & \classcell{77.85} & \classcell{73.21} & \classcell{96.71} \\
  \methodcell{CRT-Fusion-S~'24} & \thirdmetriccell{88.73} & \thirdmetriccell{55.94} & \metriccell{21.98} & \metriccell{81.40} & \classcell{13.02} & \classcell{45.16} & \classcell{38.48} & \classcell{7.34} & \classcell{18.99} & \classcell{8.90} & \classcell{63.53} & \classcell{71.99} & \classcell{74.82} & \thirdclasscell{93.38} & \classcell{84.40} & \classcell{91.19} & \classcell{86.39} & \classcell{85.53} & \thirdclasscell{98.73} \\
  \methodcell{STCOcc-S~'25} & \metriccell{78.54} & \metriccell{50.71} & \metriccell{23.48} & \metriccell{71.14} & \classcell{13.43} & \thirdclasscell{47.83} & \thirdclasscell{39.62} & \classcell{10.21} & \classcell{17.83} & \classcell{11.93} & \classcell{50.18} & \classcell{71.54} & \classcell{58.33} & \classcell{90.30} & \classcell{78.78} & \classcell{84.90} & \classcell{67.72} & \classcell{67.32} & \classcell{97.47} \\
  \methodcell{CRT-Fusion~'24} & \metriccell{80.35} & \metriccell{53.83} & \thirdmetriccell{27.02} & \metriccell{73.94} & \secondclasscell{32.58} & \classcell{46.63} & \classcell{39.49} & \secondclasscell{12.22} & \thirdclasscell{20.81} & \classcell{10.41} & \classcell{53.08} & \classcell{72.27} & \classcell{68.15} & \classcell{89.06} & \classcell{79.09} & \classcell{84.48} & \classcell{72.06} & \classcell{73.30} & \classcell{97.68} \\
  \methodcell{STCOcc~'25} & \secondmetriccell{92.54} & \secondmetriccell{60.85} & \secondmetriccell{27.66} & \secondmetriccell{85.74} & \thirdclasscell{27.13} & \secondclasscell{48.78} & \secondclasscell{44.93} & \thirdclasscell{10.23} & \bestclasscell{22.85} & \thirdclasscell{12.03} & \secondclasscell{68.84} & \classcell{70.93} & \thirdclasscell{81.60} & \secondclasscell{95.88} & \secondclasscell{89.41} & \secondclasscell{95.36} & \secondclasscell{91.85} & \secondclasscell{92.04} & \secondclasscell{99.18} \\
  \methodcell{\textbf{RoadOcc (ours)}} & \bestmetriccell{94.03} & \bestmetriccell{65.29} & \bestmetriccell{32.37} & \bestmetriccell{89.98} & \bestclasscell{37.60} & \bestclasscell{55.46} & \bestclasscell{49.36} & \bestclasscell{13.01} & \secondclasscell{22.66} & \bestclasscell{16.13} & \bestclasscell{87.66} & \classcell{72.05} & \bestclasscell{86.03} & \bestclasscell{97.85} & \bestclasscell{92.81} & \bestclasscell{96.90} & \bestclasscell{92.91} & \bestclasscell{93.66} & \bestclasscell{99.35} \\
  \bottomrule
  \end{tabular}%
  }
  \end{table}%
}
\mainresultstable
Let $\mathbf H_t^{\rm PTR,s}(\mathbf x)$ concatenate the routed multi-frame
history at scale $s$. VDSF combines this temporal evidence with
the current observation and its semantic embedding through the fusion MLP
\begin{equation}
 \mathbf q_{t,\rm fuse}^s(\mathbf x)=\mathbf q_t^s(\mathbf x)+
 \Phi_s\!\left([\mathbf q_t^s(\mathbf x),
 \mathbf H_t^{\rm PTR,s}(\mathbf x),
 \operatorname{Emb}(\boldsymbol\pi_t^s(\mathbf x))]\right).
 \label{eq:vdsf_fusion}
\end{equation}
The output residual is scattered to $\mathcal X_t^s$ and joined with the
background path. The fused volume and dynamic flow feed the next scale and are
retained for subsequent frames. The three stages use $K=[2000,512,128]$ sparse
tokens and retain $[8,4,2]$ history samples, respectively; a historical source
is valid when it has same-class support at the same height within a $3\times3$
XY neighborhood.

\subsection{Training Objective}
\label{sec:objective}

Training jointly optimizes four functional objectives.
\begin{equation}
  \mathcal L=\lambda_{\rm depth}\mathcal L_{\rm depth}+
  \mathcal L_{\rm sem}+
  \lambda_{\rm mot}\mathcal L_{\rm mot}+
  \lambda_{\rm route}\mathcal L_{\rm route}.
  \label{eq:loss}
\end{equation}
$\mathcal L_{\rm depth}$ supervises camera depth for view transformation, while
$\mathcal L_{\rm sem}$ covers full- and multi-scale occupancy together with the
static, dynamic, and DCA auxiliaries. $\mathcal L_{\rm mot}$ supervises
the dynamic flow used for backwarp sampling, whereas $\mathcal L_{\rm route}$
supervises the P/T/R router on valid dynamic voxels. Together, they separate
where history features are read from which source enters fusion. We set
$\lambda_{\rm depth}=0.5$, $\lambda_{\rm mot}=0.1$, and
$\lambda_{\rm route}=0.5$. Multi-scale semantic and route losses use relative
weights $[1,0.5,0.25,0.125]$ from native to $1/8$ resolution.

% \begin{figure}[t]
%   \centering
%   \includegraphics[width=\columnwidth]{\figdir/roadocc_sparse_selection.png}
%   \caption{\textbf{Flow-guided dynamic sparse fusion at the finest decoder
%   stage.} The visualizer reports the guidance score, selected top-$K$ mask,
%   dynamic ground truth, and semantic occupancy. VDSF reads history only at
%   selected tokens, while the flow field determines the source address of each
%   read.}
%   \label{fig:dynamic_retrieval}
% \end{figure}

\begin{figure*}[t]
  \centering
  \includegraphics[width=0.92\textwidth]{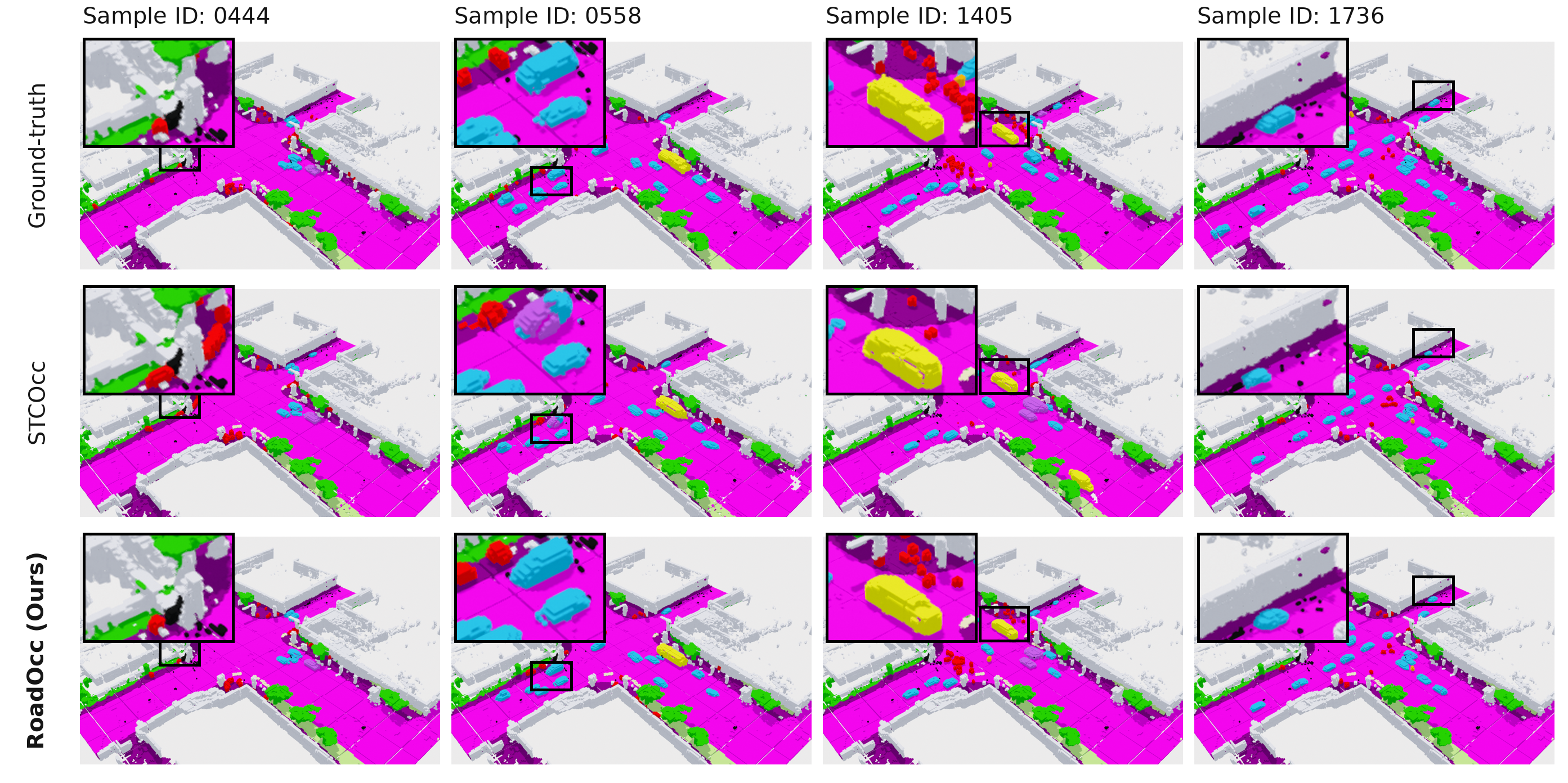}
  \caption{\textbf{Qualitative comparison with STCOcc.}
  RoadOcc more closely recovers dynamic-object presence, semantic identity,
  and geometry, including spurious pedestrian occupancy, vehicle-class
  confusion, bus-shape distortion, and an incomplete car footprint.}
  \label{fig:qualitative_comparison}
\end{figure*}
\begin{table*}[!t]
  \centering
  \caption{\textbf{Joint occupancy--velocity comparison.}}
  \label{tab:joint_results}
  \footnotesize
  \definecolor{ptrrow}{RGB}{239,247,252}
  \setlength{\tabcolsep}{1pt}
  \setlength{\belowrulesep}{0pt}
  \setlength{\aboverulesep}{0pt}
  \begin{tabular}{@{}L{3.55cm}|C{0.95cm}C{0.95cm}C{0.95cm}C{0.95cm}C{0.95cm}|C{1.90cm}|C{1.30cm}C{1.00cm}@{}}
    \toprule
    \multicolumn{1}{C{4.11cm}|}{\multirow{2}{=}{Method}} &
    \multicolumn{5}{c|}{Occupancy} & \multicolumn{1}{c|}{Velocity} &
    \multicolumn{2}{c}{Diagnostics} \\
    \cmidrule(lr){2-6}\cmidrule(lr){7-7}\cmidrule(l){8-9}
    & mIoU & Dyn. & Sta. & gIoU & {\scriptsize RayIoU} & \mbox{Direct MAVE} & \mbox{\scriptsize TP-MAVE} & DSR \\
    \midrule
    CRT-Fusion~(NIPS 2024) & 53.83 & 27.02 & 73.94 & 80.35 & 48.08 & 2.227 & 1.839 & 57.56 \\
    \quad + VDSF P/T/R & 57.15 & 28.30 & 78.79 & 83.46 & 52.34 & 1.990 & 1.958 & \underline{59.45} \\
    \rowcolor{ptrrow}\quad $\Delta$ & $+$3.32 & $+$1.28 & $+$4.85 & $+$3.11 & $+$4.26 & $-$0.237 & $+$0.119 & $+$1.89 \\
    STCOcc~(CVPR 2025) & 60.85 & 27.66 & 85.74 & 92.54 & 55.94 & 1.946 & 0.703 & 57.12 \\
    \quad + VDSF P/T/R & \underline{61.81} & \underline{28.80} & \underline{86.56} & \underline{93.59} & \underline{58.07} & \underline{1.770} & \textbf{0.648} & 58.78 \\
    \rowcolor{ptrrow}\quad $\Delta$ & $+$0.96 & $+$1.14 & $+$0.82 & $+$1.05 & $+$2.13 & $-$0.176 & $-$0.055 & $+$1.66 \\
    \midrule
    \textbf{RoadOcc (ours)} & \textbf{65.29} & \textbf{32.37} & \textbf{89.98} & \textbf{94.03} & \textbf{65.44} & \textbf{1.669} & \underline{0.661} & \textbf{61.11} \\
    \bottomrule
  \end{tabular}
\end{table*}

\section{Experiments}
\label{sec:experiments}

% \subsection{Setup}
% \label{sec:setup}

\noindent\textbf{Dataset and metrics.}
We evaluate camera-only roadside occupancy on InfraOcc
\citep{V2XReal,InfraOcc}. Four synchronized views are mapped to 18 occupancy
states. Under the InfraOcc protocol, semantic mIoU excludes construction
vehicle, trailer, other flat, and Free. Free IoU is reported separately in the
class-wise table. We report semantic and geometric IoU, next-frame dynamic IoU,
and motion coverage. Direct MAVE covers all GT dynamic voxels, while TP-MAVE
conditions on semantic matches and is paired with recall (DSR).
InfraOcc is the primary fixed-roadside protocol. We additionally evaluate
module transfer, Occ3D-nuScenes, and temporal-gap changes. The benchmark
contains 290 temporally continuous roadside sequences: 215 for training and
75 for evaluation.

\noindent\textbf{Implementation details.}
RoadOcc uses a ResNet-50 backbone, $256\times704$ images, and three aggregation
scales at 3.2, 1.6, and 0.8 m, followed by 0.4 m output voxels over
$[-64,64]\times[-64,64]\times[-4.8,1.6]$ m. Frames are
0.5 s apart. AdamW training runs for 24 epochs from $5\times10^{-4}$ with
$10^{-2}$ weight decay and cosine annealing. Comparisons share the voxel range,
ignored classes, and evaluator. Controlled variants keep the
backbone, data split, optimizer and other
configuration fixed.

% The $-$S suffix disables temporal history while
% retaining the task architecture and training protocol, whereas unmarked temporal
% variants use their corresponding memory path. 

\subsection{Main Results}
\label{sec:main_results}

% We first establish the official occupancy gain in
% Table~\ref{tab:infraocc_main_results}, then test whether this gain is accompanied
% by improved motion accuracy and historical-support coverage in
% Table~\ref{tab:joint_results}.
% Table~\ref{tab:infraocc_main_results} addresses the first using
% current-frame camera baselines. Table~\ref{tab:joint_results} addresses the
% second under a common temporal protocol with occupancy, velocity, and support
% diagnostics.

\noindent\textbf{InfraOcc occupancy benchmark.}
Table~\ref{tab:infraocc_main_results} tests whether the proposed routing chain
improves occupancy under the InfraOcc evaluator. RoadOcc achieves the highest
mIoU among the compared methods at 65.29, a $+$4.44 gain over STCOcc.
Dynamic mIoU rises by $+$4.71 to 32.37, while static mIoU rises by $+$4.24
to 89.98. Five of six dynamic classes improve, led by Bike ($+$10.47) and
Bus ($+$6.68), whereas Pedestrian changes by $-$0.19. Thus the aggregate gain covers
both traffic participants and persistent structure, with variation by class.

\noindent\textbf{Joint occupancy--velocity behavior.}
Table~\ref{tab:joint_results} evaluates motion accuracy and semantic coverage
alongside occupancy. Direct MAVE measures velocity error over all GT dynamic
voxels, while DSR measures dynamic semantic recall.
Relative to STCOcc, RoadOcc lowers Direct MAVE from 1.946 to 1.669 and raises
DSR from 57.12 to 61.11, improving motion accuracy alongside dynamic recovery.
Adding VDSF P/T/R to CRT-Fusion and STCOcc likewise raises Dyn. by $+$1.28 and
$+$1.14, respectively, while improving Direct MAVE and DSR. The consistent
trend supports the module's utility beyond the full RoadOcc architecture.
TP-MAVE conditions on semantic matches and must be
read with DSR. For CRT-Fusion, the $+$1.89 DSR gain indicates broader dynamic
recovery even though the conditional error over matched voxels increases.

\noindent\textbf{Qualitative occupancy recovery.}
Figure~\ref{fig:qualitative_comparison} supports the aggregate dynamic gain with
representative matched scenes. STCOcc produces
stale or spurious pedestrian occupancy, confuses a vehicle class, distorts a
bus footprint, and incompletely recovers a car. RoadOcc more closely recovers
the GT class and extent in these examples while preserving the roadway.

\begin{wraptable}[10]{r}{0.50\columnwidth}
  \vspace{-5pt}
  \centering
  \captionof{table}{\textbf{Component ablation.}}
  \label{tab:temporal_design}
  \footnotesize
  \renewcommand{\arraystretch}{1.05}
  \setlength{\tabcolsep}{0pt}
  \setlength{\belowrulesep}{0pt}
  \setlength{\aboverulesep}{0pt}
  \begin{tabular}{@{}C{1.00cm}|C{0.82cm}C{0.82cm}C{0.82cm}C{0.82cm}|C{0.85cm}C{0.85cm}C{0.85cm}@{}}
    \toprule[1.0pt]
    \multicolumn{1}{C{0.95cm}|}{\multirow{2}{=}{Base}} &
    \multicolumn{4}{c|}{Design} &
    \multicolumn{3}{c}{Output $\uparrow$} \\
    \cmidrule(lr){2-5}\cmidrule(l){6-8}
     & DCA & VVE & VDSF & P/T/R & All & Dyn. & Nxt. \\
    \midrule
    \cmark &  &  &  &  & 60.85 & 27.66 & 26.74 \\
    \cmark &  &  & \cmark &  & 61.05 & 28.01 & 26.99 \\
    \cmark & \cmark &  &  &  & 61.69 & 28.55 & 27.64 \\
    \cmark & \cmark &  & \cmark &  & 63.01 & 30.10 & 29.44 \\
    \cmark & \cmark & \cmark & \cmark &  & 64.17 & 30.97 & 30.78 \\
    \cmark & \cmark & \cmark & \cmark & \cmark & \textbf{65.29} & \textbf{32.37} & \textbf{31.72} \\
    \bottomrule[1.0pt]
  \end{tabular}
\end{wraptable}

\noindent\textbf{Transfer and temporal gaps.}
On mask-trained Occ3D-nuScenes, RoadOcc reaches 45.01 overall and 39.80 dynamic
mIoU, improving over STCOcc by 0.41 and 0.79 points, respectively. Without
fine-tuning the $0.5$ s-trained checkpoints, testing at $1.5$ s yields 33.49
Dyn. for RoadOcc and 28.58 for STCOcc on the same 1,520 anchors. These results extend the empirical support
beyond the primary dataset and training cadence. They evaluate the complete
model rather than isolate P/T/R.

\subsection{Ablation Studies}
\label{sec:ablation}

Using \citet{STCOcc} as the baseline, we hold the backbone, token budget, temporal-memory
configuration, and training protocol fixed unless explicitly varied.
Tables~\ref{tab:temporal_design}, \ref{tab:flow_branch}, and
\ref{tab:motion_counterfactual} then examine component composition,
historical-address correction, and P/T/R source selection, respectively.

\begin{figure*}[t]
  \centering
  \includegraphics[width=1.0\textwidth]{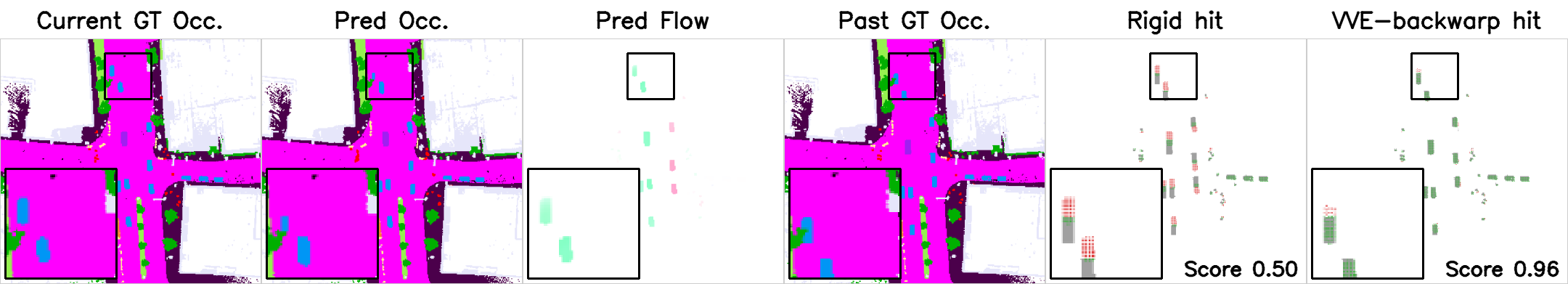}
  \caption{\textbf{Velocity-guided historical retrieval.} With queried
  dynamic targets fixed by current GT, the panels show the end-to-end occupancy
  prediction and compare a fixed-coordinate read with the VVE backtrace into aligned past
  GT. Black boxes are enlarged in the insets. \hitbox{hitmatch} same-class
  matches, \hitbox{hitmismatch} unmatched targets, and \hitbox{hitother} other
  queried voxels are shown in green, red, and gray, respectively.
  The score is the same-class hit rate within the one-voxel matching radius.}
  \label{fig:vve_backwarp}
\end{figure*}

\noindent\textbf{Overall ablation.}
Table~\ref{tab:temporal_design} follows the routing chain. VDSF alone adds only
0.20 mIoU. Adding VDSF to the DCA variant raises Dyn. by $+$1.55 and Nxt.
by $+$1.80. VVE then adds $+$1.16 mIoU, and P/T/R adds a further $+$1.12
mIoU. This progression supports combining candidate selection, motion-based
retrieval, and source routing. Tables below examine latter two design in more detail.

\noindent\textbf{Address correction and source selection.}
With capacity and token budget fixed, the fixed-coordinate and VVE reads in
Table~\ref{tab:flow_branch} differ only in historical address. VVE raises Dyn. by
$+$0.87, lowers dMAVE from 2.146 to 1.690, and raises DSR by $+$1.51.
P/T/R then adds $+$1.40 Dyn. and $+$1.76 DSR, showing that learned source
selection improves on motion-corrected retrieval.

\begin{wraptable}[18]{r}{0.48\columnwidth}
  \centering
  \captionof{table}{\textbf{Controlled address and source selection.} All
  variants use the same sparse-token budget and model capacity.}
  \label{tab:flow_branch}
  \footnotesize
  \renewcommand{\arraystretch}{1.05}
  \setlength{\tabcolsep}{0.4pt}
  \setlength{\belowrulesep}{0pt}
  \setlength{\aboverulesep}{0pt}
  \begin{tabular}{@{}C{1.50cm}|C{1.00cm}C{1.00cm}|C{0.75cm}C{0.75cm}C{0.75cm}C{0.75cm}@{}}
    \toprule[1.0pt]
    \multicolumn{1}{>{\centering\arraybackslash}m{1.38cm}|}{\multirow{2}{*}{Variant}} &
     \multicolumn{2}{c|}{Design} & \multicolumn{4}{c}{Output} \\
    \cmidrule(lr){2-3}\cmidrule(l){4-7}
    & VVE & P/T/R & {\scriptsize Dyn.$\uparrow$} & {\scriptsize Nxt.$\uparrow$} & {\scriptsize dM$\downarrow$} & {\scriptsize DSR$\uparrow$} \\
    \midrule
    \mbox{Fixed read} &  &  & 30.10 & 29.44 & 2.146 & 57.84 \\
    \mbox{VVE read} & \cmark &  & 30.97 & 30.78 & 1.690 & 59.35 \\
    \textbf{Full} & \cmark & \cmark & \textbf{32.37} & \textbf{31.72} & \textbf{1.669} & \textbf{61.11} \\
    \bottomrule[1.0pt]
  \end{tabular}
  \centering
  \captionof{table}{\textbf{P/T/R routing controls.} Mean $\pm$ standard
  deviation is reported for the multi-seed variants. w/o R denotes full w/o Refresh.}
  \label{tab:motion_counterfactual}
  \footnotesize
  \renewcommand{\arraystretch}{1.05}
  \setlength{\tabcolsep}{0.0pt}
  \setlength{\belowrulesep}{0pt}
  \setlength{\aboverulesep}{0pt}
  \begin{tabular}{@{}L{1.10cm}|C{1.40cm}C{1.40cm}C{1.40cm}C{1.40cm}@{}}
    \toprule[1.0pt]
    Variant & Dyn.$\uparrow$ & Nxt.$\uparrow$ & DSR$\uparrow$ & dM$\downarrow$ \\
    \midrule
    None & 30.97{\scriptsize$\pm$.27} & 30.78{\scriptsize$\pm$.34} & 59.35{\scriptsize$\pm$.39} & 1.690{\scriptsize$\pm$.019} \\
    Gate & 31.25{\scriptsize$\pm$.20} & 30.94{\scriptsize$\pm$.30} & 59.75{\scriptsize$\pm$.45} & 1.685{\scriptsize$\pm$.018} \\
    State & 31.88{\scriptsize$\pm$.18} & 31.38{\scriptsize$\pm$.25} & 60.52{\scriptsize$\pm$.40} & 1.675{\scriptsize$\pm$.017} \\
    w/o R & 32.05{\scriptsize$\pm$.12} & 31.41{\scriptsize$\pm$.18} & 60.69{\scriptsize$\pm$.05} & 1.679{\scriptsize$\pm$.033} \\
    \textbf{Full} & \textbf{32.37{\scriptsize$\pm$.11}} & \textbf{31.72{\scriptsize$\pm$.26}} & \textbf{61.11{\scriptsize$\pm$.38}} & \textbf{1.669{\scriptsize$\pm$.020}} \\
    \bottomrule[1.0pt]
  \end{tabular}
\end{wraptable}

\noindent\textbf{P/T/R routing controls.}
Table~\ref{tab:motion_counterfactual} compares routing variants within
DCA--VVE--VDSF. None omits P/T/R. Gate adaptively mixes the same candidates.
State adds P/T/R supervision without executing its probabilities. Full uses
the supervised routes in fusion. Full w/o R masks $\alpha_R$ and renormalizes
$(\alpha_P,\alpha_T)$. The complete control contract is given in the appendix.
Across three seeds, Full reaches 32.37 $\pm$ 0.11 Dyn., compared with
30.97 $\pm$ 0.27 for None and 31.25 $\pm$ 0.20 for Gate. State and Full improve
successively by 0.63 and 0.49 points over the preceding variant. The clearest
gain is in dynamic occupancy. The dMAVE change from None to Full is smaller
(0.021 m/s), with standard deviations of 0.019 and 0.020 m/s.
Removing Refresh from Full reduces Dyn., Nxt., and DSR by 0.32, 0.31, and
0.42 points, respectively, while increasing dMAVE by 0.010 m/s. Under the same
three-seed protocol, the 0.32-point Dyn. gap exceeds either reported run-to-run
standard deviation (0.11 and 0.12). We treat the other metric changes as
supporting trends rather than standalone significance claims.
% \noindent\textbf{Dynamic behavior.}
% Recall and footprint diagnostics distinguish missing support from semantic
% confusion, while ghost occupancy measures stale evidence left at a previous
% location. Together with the speed-stratified analysis in
% Table~\ref{tab:temporal_verification}, these diagnostics test whether the model
% improves moving objects without disturbing already aligned slow or stationary
% participants.

\noindent\textbf{Routing scope.}
Equation~\ref{eq:ptr_target} supervises motion-conditioned semantic preferences,
not instance correspondence. Labels use the nearest history, and each scale
shares its predicted route across valid memory slots. GT-conditioned routing
visualizations separate source choices from occupancy detection. They do not
measure candidate misses or guarantee support at every predicted historical
address. Dyn., Nxt., Direct MAVE, and DSR provide complementary end-to-end metrics.

\begin{wrapfigure}{r}{0.68\columnwidth}
  \vspace{-7pt}
  \centering
  \includegraphics[width=\linewidth]{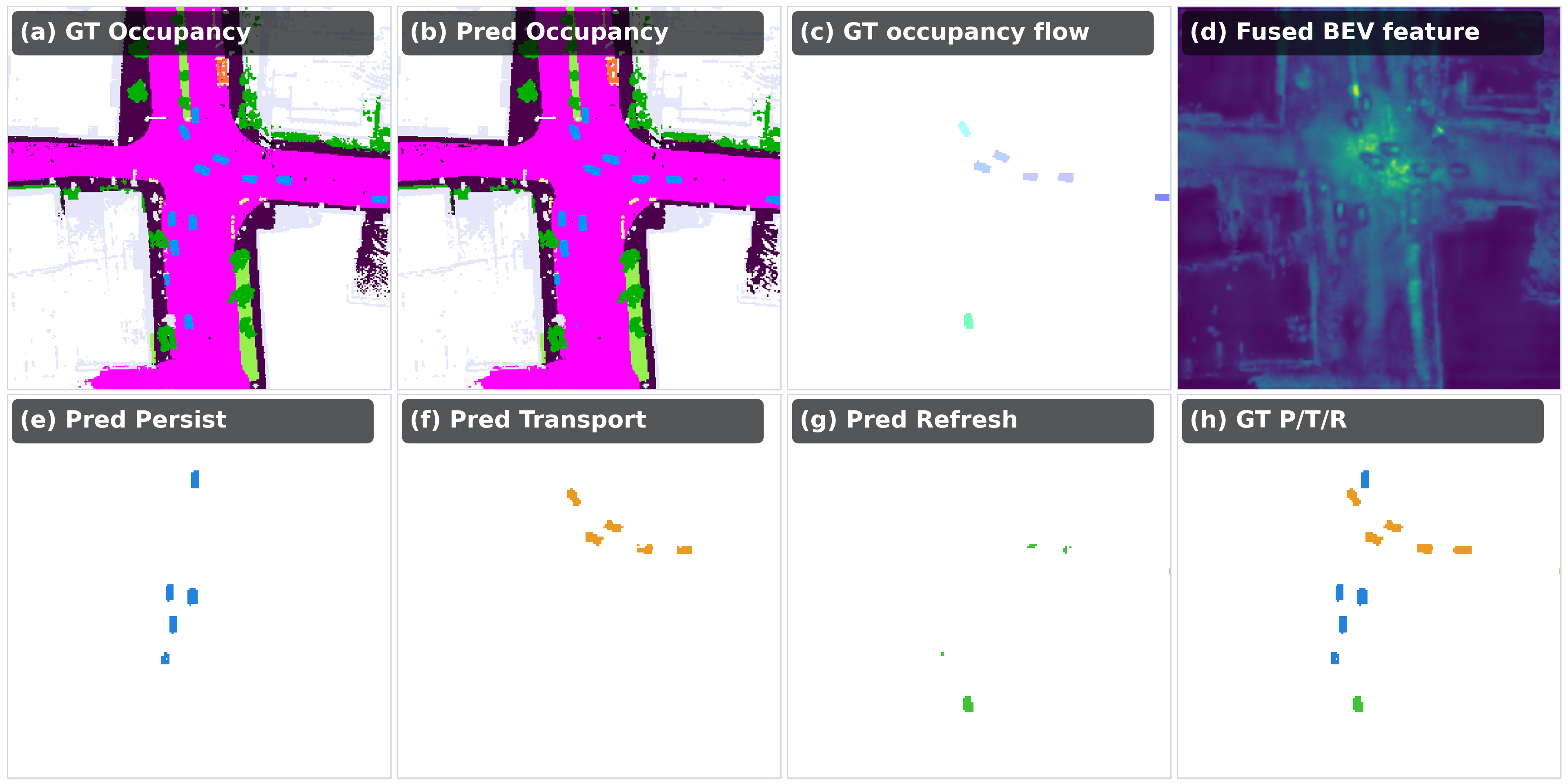}
  \vspace{-15pt}
  \caption{\textbf{Predicted P/T/R on GT dynamic support.}}
  \vspace{-10pt}
  % Current GT/predicted
  % occupancy, GT flow, and fused BEV features are paired with Persist,
  % Transport, Refresh, and combined routing targets.
  \label{fig:ptr_control}
\end{wrapfigure}
\noindent\textbf{Visual analysis.}
Figure~\ref{fig:vve_backwarp} compares zero-motion reading with VVE backtracing.
Current GT fixes the queried dynamic voxels and their classes, while predicted
occupancy provides the end-to-end context. Red-to-green changes show dynamic
flow relocating queries through backwarp sampling onto same-class past GT,
isolating address quality from the subsequent routing decision. On the same GT
dynamic support, Figure~\ref{fig:ptr_control} then shows Persist following
fixed-coordinate support, Transport following displacement, and Refresh favoring
current evidence. Together, the visualizations inspect historical addresses
and learned source preferences on dynamic targets.

\section{Conclusion}
\label{sec:conclusion}

In this work, we present RoadOcc, a voxel-level evidence-source routing framework for temporal
roadside occupancy. Beyond implicit temporal aggregation,
RoadOcc explicitly learns where history should be retrieved and
how strongly it should be reused. The proposed DCA uses current image evidence to identify
motion-sensitive voxels, while VVE estimates their historical source addresses, and
VDSF combines evidence through Persist, Transport, or Refresh routing. On
the InfraOcc protocol, the unified formulation achieves the highest occupancy
accuracy among the compared methods while improving dynamic recovery and velocity estimation. Extensive experiments
further demonstrate the superiority of our design.
Future work will extend this principle to broader moving-platform settings, longer temporal horizons, and
instance-level modeling.
% Cross-dataset validation is currently constrained by the absence of another
% compatible real-world benchmark for dense fixed-roadside occupancy. 
% These results support explicit physical
% provenance as a coherent and interpretable complement to conventional temporal
% fusion.
\section*{AI Use Statement}
Generative AI tools were used for literature organization, reviewer-style
feedback, and language editing. The authors verified all citations, technical
descriptions, analyses, and revisions, and take responsibility for the final
content of this work.
\bibliography{iclr2027}
\bibliographystyle{iclr2027}
% Keep the supplementary material in the same single-column ICLR document,
% after the main-paper references.
\end{document}